\documentclass[runningheads]{llncs}
\usepackage[T1]{fontenc}
\usepackage{graphicx}
\usepackage{amsmath}
\usepackage{booktabs}
\usepackage{float}
\usepackage{makecell}
\usepackage{booktabs}
\usepackage{url}        
\usepackage[colorlinks=true, linkcolor=blue, urlcolor=blue]{hyperref}

\begin{document}
\title{MIRAGE: Retrieval and Generation of Multimodal Images and Texts for Medical Education}
\titlerunning{MIRAGE: A Tool for Medical Education}
%
%

\author{Miguel Díaz Benito\inst{1}\orcidID{0009-0002-2110-9972} \and
Cecilia Diana-Albelda\inst{1}\orcidID{0009-0009-9210-0853} \and
Álvaro García-Martín\inst{1}\orcidID{0000-0002-1705-3972} \and
Jesus Bescos\inst{1}\orcidID{0000-0001-6238-6859} \and
Marcos Escudero Viñolo\inst{1}\orcidID{0000-0002-9156-3428} \and
Juan Carlos SanMiguel \inst{1}\orcidID{0000-0002-4999-2851}}

\authorrunning{M. Díaz Benito et al.}

\institute{Universidad Autónoma de Madrid, 28049 Madrid, Spain\\
\email{miguel.diazbenito@estudiante.uam.es}}

%
\maketitle              

\begin{abstract}
Access to diverse, well-annotated medical images with interactive learning tools is fundamental for training practitioners in medicine and related fields to improve their diagnostic skills and understanding of anatomical structures. While medical atlases are valuable, they are often impractical due to their size and lack of interactivity, whereas online image search may provide mislabeled or incomplete material. To address this, we propose MIRAGE, a multimodal medical text and image retrieval and generation system that allows users to find and generate clinically relevant images from trustworthy sources by mapping both text and images to a shared latent space, enabling semantically meaningful queries. The system is based on a fine-tuned medical version of CLIP (MedICaT-ROCO), trained with the ROCO dataset, obtained from PubMed Central. MIRAGE allows users to give prompts to retrieve images, generate synthetic ones through a medical diffusion model (Prompt2MedImage) and receive enriched descriptions from a large language model (Dolly-v2-3b). It also supports a dual search option, enabling the visual comparison of different medical conditions. A key advantage of the system is that it relies entirely on publicly available pretrained models, ensuring reproducibility and accessibility. Our goal is to provide a free, transparent and easy-to-use didactic tool for medical students, especially those without programming skills. The system features an interface that enables interactive and personalized visual learning through medical image retrieval and generation. The system is accessible to medical students worldwide without requiring local computational resources or technical expertise, and is currently deployed on Kaggle: \color{blue}\href{http://www-vpu.eps.uam.es/mirage}{http://www-vpu.eps.uam.es/mirage}\color{black}


\end{abstract}
\vspace{-0.1in}
\section{Introduction}
\vspace{-0.05in}
We live in an era of rapid transformation driven by advances in science, technology and artificial intelligence. In this context, educating the next generation of scientists requires equipping them with tools that are not only effective but also trustworthy and accessible. Traditionally, medical students have relied on atlases to study anatomical and pathological structures. Internet search engines offer rapid access to visual information, but often at the cost of reliability due to lack of context, labeling, or source verification  \cite{misinfo}.

The emergence of Large Language Models (LLMs) \cite{SurveyLLM} has revolutionized access to textual information. Students can now obtain detailed answers to complex medical questions almost instantly. Yet, many responses are not grounded in validated medical data \cite{Haluc}. Previous studies have combined multimodal learning with healthcare applications \cite{Multimodal}. For instance, MedCLIP \cite{MedCLIP}, an adaptation of CLIP \cite{CLIP} for medical imaging, achieved significant improvements in zero-shot classification and retrieval. However, LLMs such as BioGPT \cite{BioGPT} and Dolly-v2 \cite{DollyV2} have been used to enrich text, generating more informative and context-aware descriptions. This is particularly important in the medical field, where such descriptions are needed quickly to support fast and accurate diagnosis.

Given their role in education and diagnosis, medical images demand systems that combine AI flexibility with clinical rigor. Recent studies have addressed this need, with image retrieval gaining increasing attention \cite{ImageRetrieval}. This process usually relies on multimodal learning, where models are trained to align visual and textual information within a shared latent space, enabling semantically meaningful comparisons across modalities. In the biomedical field, this is useful for various purposes such as clinical decision support, medical education and research \cite{MedicalImUses}.

Another growing area in computer vision is image generation, with diffusion models becoming the state of the art \cite{DiffModel}. This is especially relevant in medicine, where privacy and consent issues limit real data availability. In such cases, synthetic data offers a practical and ethical alternative \cite{DiffusionModelsMedImage}, particularly for rare or underrepresented conditions. In learning contexts, it also enables simplified and user-tailored representations of medical concepts.

Most generative AI tools in medical education focus on text, with limited integration of imaging. Prior work has applied AI to self-directed learning and tutoring \cite{Preiksaitis2023,Boscardin2024,Eysenbach2023,Stretton2024,Rao2025,Janumpally2025}, but remains largely text-centric. Clinical use is often narrow and lacks image-based training. In contrast, our system unifies retrieval, comparison, description, and generation of medical images, offering a richer multimodal learning experience. To our knowledge, no publicly available tool integrates all these components, limiting practical use in education. 

In this paper we propose a multimodal medical image retrieval and generation system that enhances learning by integrating real image retrieval, synthetic image generation, and natural language description. Built upon the ROCO (Radiology Objects in COntext) dataset \cite{ROCO}, compiled from PubMed, our system retrieves the most semantically relevant medical image from a user prompt, returning it along with a description and a synthetic image generated from the input. Furthermore, it generates synthetic images using a diffusion-based model trained on medical data and provides enriched textual descriptions through an LLM. Additionally, the system supports dual-query functionality, enabling users to visually compare and contrast different concepts, by applying latent space arithmetic to the textual embeddings. Specifically, a query embedding is computed by subtracting the embedding of one concept and adding that of another, allowing MIRAGE to retrieve and generate results that reflect the semantic shift.

This paper makes three main contributions. First, we present a unified multimodal system that combines medical image and text retrieval, concept-level comparison via latent space manipulation, and synthetic image generation from user prompts. Second, the entire pipeline relies exclusively on publicly available pretrained models, ensuring reproducibility. Third, the system is deployed on Kaggle with user-friendly instructions, making it accessible to students without programming skills. Additionally, we validate its semantic consistency through both quantitative and qualitative analyses in realistic educational scenarios.

\vspace{-0.05in}
\section{Methodology: MIRAGE Design and Deployment}\label{Methodology}
\begin{figure}[h]
    \centering
    \includegraphics[width=1\linewidth]{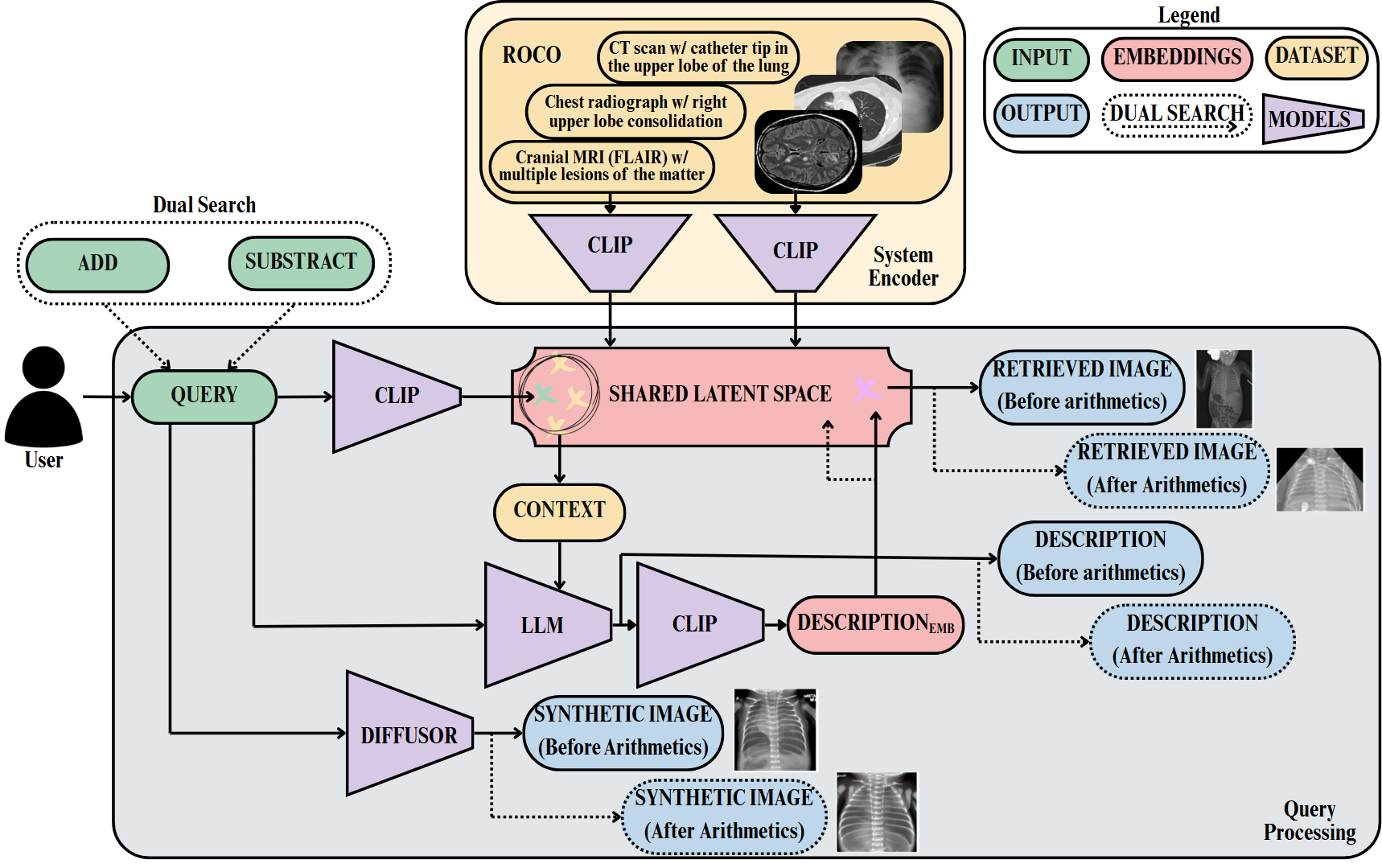}
    \caption{Overview of MIRAGE. A user query is encoded and compared to the ROCO dataset in a shared latent space to retrieve the most relevant image, generate an enriched description, and synthesize a new image. The system also supports dual-concept comparison using latent space arithmetic.}
    \label{fig:pipeline}
\end{figure}

\vspace{-0.05in}
This section presents the methodology behind MIRAGE, which consists of three modules: the dataset acts as a reference atlas; images and captions are embedded into a shared latent space; and user queries return a real image, an enriched description, and a synthetic image. Figure \ref{fig:pipeline} illustrates the steps of the pipeline. 

\subsection{System encoder: multimodal embedding}
\vspace{-0.025in}
To enable multimodal retrieval, we import both images and captions from the ROCO dataset via a CLIP-based model fine-tuned for medical content, CLIP-ViT-L-14-448px-MedICaT-ROCO.\footnote{Model publicly available at \url{https://huggingface.co/luhuitong/CLIP-ViT-L-14-448px-MedICaT-ROCO}} 
The model embeds visual and textual data into a shared latent space, enabling semantic comparison between queries and medical images. Captions and images are encoded via text and visual encoders, producing normalized embeddings stored for fast cosine similarity retrieval \cite{cosinesim}. 

\vspace{-0.1in}
\subsection{Query processing: multimodal retrieval}
\vspace{-0.025in}
We describe the pipeline workflow to obtain an output including an original image from the ROCO dataset, a synthetic image, and a description. If the user specifies an element to subtract and another to add, MIRAGE performs a conceptual modification by retrieving the most similar image from ROCO that reflects the change. It also generates a synthetic image and a revised description. 

\vspace{-0.2in}
\subsubsection{Retrieval} \label{retrieval}
Our retrieval pipeline is designed to provide not only images, but also medical descriptions for educational purposes. When a user submits a query describing a condition or anatomical structure, the system first encodes the query into an embedding using a CLIP model fine-tuned on medical content.

This embedding is compared against the entire database of precomputed image embeddings using cosine similarity. The system retrieves the top-\textit{k} most similar captions, which are then used as contextual input for a general-purpose instruction-following language model, Dolly-v2-3b \cite{DollyV2}. The model generates an enriched caption, offering a description of the queried concept.

Next, the generated caption is re-encoded using CLIP, and a second retrieval compares it against all precomputed image embeddings. The system then returns the most similar image along with the enriched description from Dolly.

\vspace{-0.1in}
\subsubsection{Dual Search and Latent Comparison}
Distinguishing visually similar but clinically distinct conditions is a common challenge in medical education \cite{LearningSimilarThings}. To address this, our system includes a dual search module allowing users to compare two medical concepts side by side. This uses latent space arithmetic \cite{Arithm1,Arithm2}, which modifies text features directly in the shared latent space. 

\begin{figure}[H]
    \centering
    \includegraphics[width=0.6\linewidth]{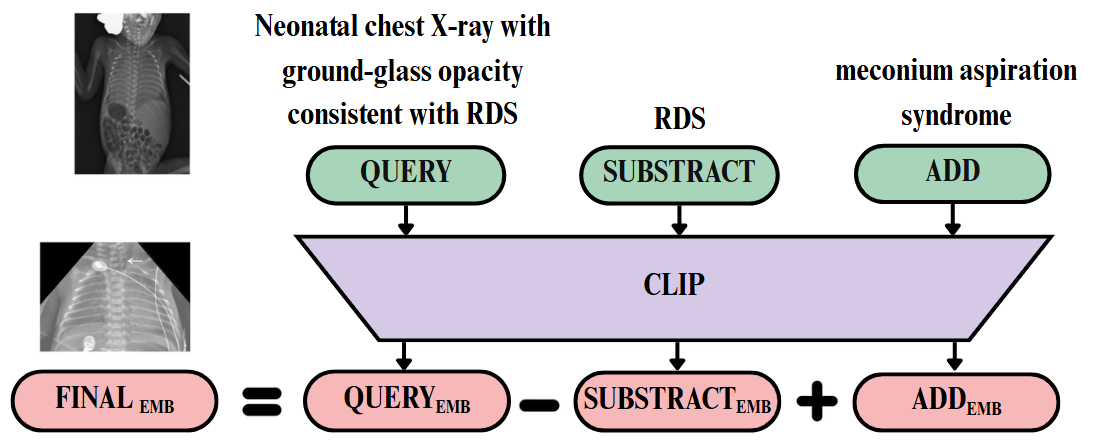}
    \caption{Dual search illustration. The system combines three text inputs, Query, Subtract, and Add, to generate a new embedding that reflects the intended concept shift.}
    \label{fig:DualSearch}
\end{figure}

As shown in Figure \ref{fig:DualSearch}, when the users want to explore the difference between two related conditions, they provide a base query and specify two terms: one representing a concept to remove, and another to introduce. All textual inputs are embedded into a shared semantic space using the same CLIP model that underlies the rest of the system. Then, it creates a new query embedding by subtracting one concept and adding the other as shown in Equation \ref{eq}.
\vspace{-0.025in}
\begin{equation}
\textit{embedding}_{\text{modified}} = \textit{embedding}_{\text{original}} 
- \textit{embedding}_{\text{term}_{\text{subtract}}}
+ \textit{embedding}_{\text{term}_{\text{add}}}
\label{eq}
\end{equation}

Here, $\text{embedding}_{\text{original}}$ represents the embedding of the initial query of the user,  
$\text{embedding}_{\text{term}_\text{subtract}}$ corresponds to the concept the user wants to remove (e.g., benign), and  
$\text{embedding}_{\text{term}_\text{add}}$ represents the concept the user wants to introduce (e.g., malignant).  
The resulting $\text{embedding}_{\text{modified}}$ reflects a semantic transformation of the original query toward the new concept. Both the original and modified embeddings are then used to retrieve their respective top-1 most similar images from the database, enabling visual comparison.

\vspace{-0.2in}
\vspace{-0.25in}

The system is designed with human-AI collaboration in mind. The dual search feature enables students to interactively explore clinical hypotheses by comparing conditions, while the descriptions provide educational context that complements visual learning. This encourages active learning rather than passive image browsing.

\vspace{-0.1in}
\subsubsection{Synthetic Image Generation}
In addition to retrieving real medical images, the system can generate synthetic ones from prompts. Comparing both helps highlight key concepts in the generated image, aiding interpretation and concept-based learning. We use Prompt2MedImage\footnote{Available for free at \url{https://huggingface.co/Nihirc/Prompt2MedImage}.}\cite{Prompt2MedImageExample}, a diffusion-based model trained for medical content. It handles medical prompts and can produce synthetic images across different modalities (e.g., MRI, CT, X‑ray), leveraging a latent diffusion architecture and a fixed CLIP ViT‑L/14 text encoder\cite{ViT}.

MIRAGE uses the original prompt provided by the user when generating the synthetic image. While an enriched description is produced from the initial query (see Section \ref{retrieval}), the original prompt is used for the image generation step, as longer descriptions do not lead to better synthetic data 


\vspace{-0.05in}
\section{Evaluation Methodology}
\vspace{-0.05in}
In this section, we present ROCO, the dataset used in this work, along with the experimental setup and the evaluation metrics employed to assess the system.

\vspace{-0.1in}
\subsection{Setup}
\vspace{-0.025in}
All components of MIRAGE, except for the LLM, are pretrained on the ROCO dataset \cite{ROCO}, a large-scale collection of medical images designed for multimodal research. ROCO consists of over 81,000 images extracted from the Open Access subset of PubMed Central, with a wide variety of imaging modalities, including X-ray, CT and MRI among others. Each image is accompanied by a caption written by medical experts, typically extracted from figure legends or article content. These captions usually indicate the image type and the pathology depicted.

The system runs on the cloud infrastructure of Kaggle, leveraging their free GPU tier (NVIDIA T4 GPUs). The users can access MIRAGE without any installation. The system typically processes queries in 40 to 60 seconds, making it suitable for interactive learning. By designing the system to run entirely within this environment, we ensure that all experiments are reproducible and aligned with the principles of accessibility and open science. In addition, detailed setup instructions are provided for users with no programming experience.

\vspace{-0.1in}
\subsection{Semantic Consistency and System Evaluation}
\vspace{-0.025in}
To assess the effectiveness and semantic coherence of our multimodal search engine, we propose an evaluation strategy combining quantitative embedding analysis and qualitative visual inspection.

Firstly, we evaluate the consistency of the embedding space by computing cosine similarity scores across different types of input pairs: medical text prompts (similar and dissimilar) and image-caption pairs. The aim was to assess whether the CLIP-based encoder preserves meaningful clinical relationships in the latent space. In the text-only evaluation, we compared 100 prompt pairs (50 similar, 50 dissimilar) to verify that semantically related expressions (e.g., synonyms for the same condition) yield higher similarity scores than unrelated ones. Based on these results, we established a threshold $th_{\text{CC}}$ to separate similar from dissimilar pairs and computed classification accuracy. For multimodal evaluation, we applied the same procedure to 50 real image-caption pairs and 50 randomly mismatched ones, calculating an optimal threshold $th_{\text{IC}}$ and the corresponding classification accuracy. Finally, to assess the alignment of synthetic images, we repeat the previous experiment using synthetic images generated from the dataset captions as prompts, computing the cosine similarity between each caption and its corresponding synthetic image. We then estimate a threshold to separate true caption–image pairs from randomly mismatched ones, evaluating whether the model can reliably distinguish between them.  

Secondly, we conducted a qualitative evaluation using example queries representative of real-world medical learning scenarios. For each case, we present the top-retrieved real image, the enriched caption generated by the LLM, and the synthetic image produced by the diffusion model. In cases involving dual search, we used latent space arithmetic to modify the original query and analyzed how this transformation influenced both retrieval and generation results.
 
\vspace{-0.05in}
\section{Results}\label{Results}
\vspace{-0.05in}
We evaluate our system through qualitative examples and quantitative semantic similarity analysis, aiming to assess the coherence and usefulness of the retrieved and generated outputs, particularly in medical education contexts.

\subsection{Semantic and Multimodal Embedding Analysis}
\vspace{-0.025in}

To assess the consistency and explainability of the system, we evaluate the embedding space across textual and multimodal dimensions by measuring cosine similarity between pairs of medical prompts, real image–caption pairs, and synthetic image–caption pairs. Semantically similar inputs consistently yield higher similarity scores than dissimilar ones across all setups, with classification accuracies close to or above 97\%. Results are summarized in Table \ref{tab:similarity-results}, and suggest that the model is able to link similar texts and images.

\vspace{-0.1in}
\begin{table}[h]
\centering
\begin{tabular}{lccc}
\toprule
\textbf{Evaluation Type} & \makecell{\textbf{Mean Similarity}\\ \textbf{(Similar / Dissimilar)}} & \textbf{Threshold} & \textbf{Accuracy} \\
\midrule
Caption–Caption           & $0.770 \pm 0.079$ / $0.394 \pm 0.063$   & $0.582$ & $99\%$ \\
Image–Caption (Real)      & $0.386 \pm 0.022$ / $0.086 \pm 0.0922$  & $0.236$ & $97\%$ \\
Image–Caption (Synthetic) & $0.287 \pm 0.056$ / $0.074 \pm 0.082$   & $0.230$ & $97\%$ \\
\bottomrule
\end{tabular}
\vspace{2mm}  
\caption{Cosine similarity scores and classification accuracy for text–text, image–text, and synthetic image–text alignment.}
\label{tab:similarity-results}
\end{table}

\vspace{-0.5in}
\subsection{Qualitative Examples}
\begin{figure}[H]
    \centering
    \includegraphics[width=1.0\linewidth]{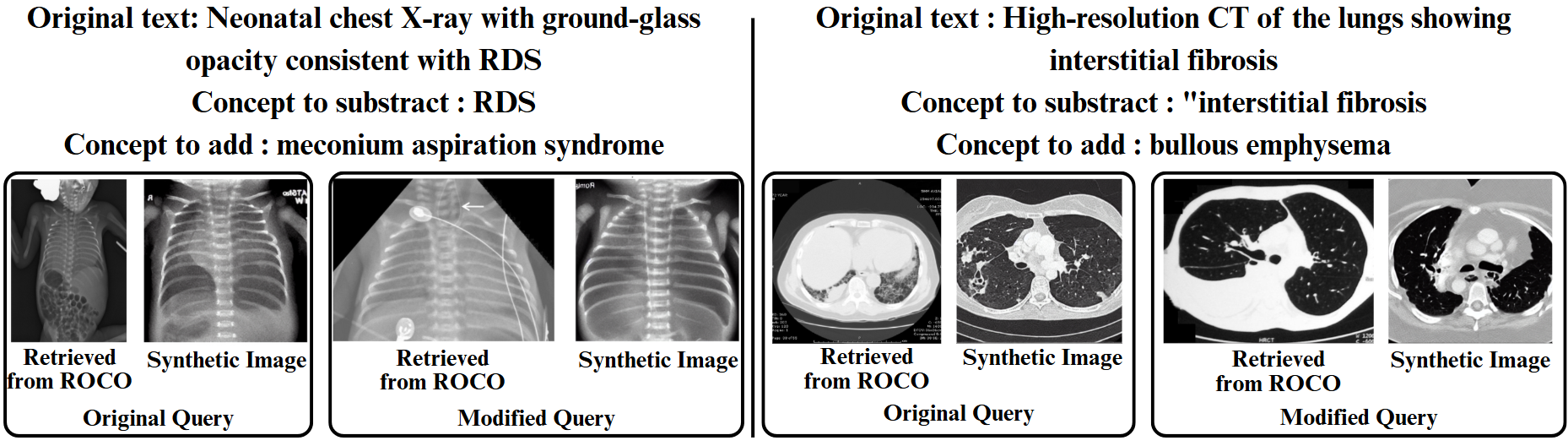}
    \caption{
Dual search comparison for two clinical scenarios. In the first case (left), the prompt “RDS” is modified to “Meconium aspiration syndrome (MAS)”. In the second case (right), “interstitial fibrosis” is replaced by “bullous emphysema”. Each example displays retrieved and synthetic images before and after the concept substitution, illustrating the semantic and visual shift induced by the latent manipulation.}
    \label{fig:neonatal}
\end{figure}

\vspace{-0.025in}
To demonstrate the capabilities of  the system, we present a medical query and show the retrieved image from the ROCO dataset, and a synthetic image produced by the Prompt2MedImage diffusion model, along with a real image and synthetic image generated after the substituted concept to highlight differences. Figure \ref{fig:neonatal} shows a sample case using the query “Neonatal chest X-ray with ground-glass opacity consistent with RDS”. For clarity and space, the enriched description is omitted from the Figure; additional examples with full captions and images are available in the Kaggle repository. MIRAGE retrieves real images visually and semantically aligned with the user query, while synthetic images capture key anatomical and pathological features from the text.

\vspace{-0.2in}

\vspace{-0.1in}
\section{Conclusions}\label{Conclussions}
\vspace{-0.05in}
In this work, we present an educational tool designed to help medical students explore and understand medical images more effectively. The system combines image retrieval, language models, and AI-generated visual content. It is based on the ROCO dataset and uses a CLIP model fine-tuned for medical data to link text and images in a shared embedding space. This advances in democratizing medical AI education by utilizing free cloud resources and pretrained models, eliminating common barriers such as computational requirements, specialized software, and technical expertise. This accessibility is particularly important for students in developing regions or institutions with limited IT infrastructure.

A key feature of the system is its ability to compare two medical concepts side by side through a dual search. By adjusting the prompt, users can visualize how different but related conditions appear, both visually and semantically. Our analysis demonstrates that the model groups similar concepts together and separates unrelated ones in a meaningful way.

We evaluated the system using cosine similarity analysis. The results show that the model maintains a consistent alignment between text and image representations, even with limited data. Tests with random prompts confirmed that the model can clearly distinguish between unrelated inputs.

There are still some limitations. Alignment between image and text is moderate in some cases, especially with synthetic content. The current version uses only about 5\% of the ROCO dataset due to computational limits. With more powerful hardware, indexing a larger portion could improve variety and accuracy.
 
Overall, this tool not only helps users find and understand medical images more easily, but also allows them to compare related medical concepts and learn visually through images generated by AI. We believe this tool can be especially helpful in educational settings, particularly for students who are not yet familiar with using complex medical atlases. To make the system freely accessible, we deployed the entire pipeline on Kaggle, which offers a free GPU environment. This allows anyone to use the system, run searches, and view results directly in the browser without needing to install anything or have a powerful computer.

In the future, we plan to add the option to execute only selected parts of the pipeline, depending on the user's preferences. Additionally, we aim to develop a user-friendly web interface. While Kaggle is indeed a convenient platform, we believe that a dedicated website and integration with existing medical education platforms could further simplify and enhance the user experience. Finally, we aim to validate our system,  immediately deployable via Kaggle, through user studies involving medical practitioners and students in related fields.

\textbf{Prospect of application.}
Our multimodal system retrieves relevant medical images, provides enriched description, enables concept comparison through dual search, and generates synthetic images. Integrated into medical education platforms, it supports interactive, concept-driven learning, aids clinical reasoning, and enhances the visual‑textual understanding of the students, making advanced AI accessible to students worldwide without technical barriers.

\textbf{Disclosure of Interests}
The authors declare no competing interests.

\textbf{Acknowledgements}
This work has been partially supported by the Ministerio de Ciencia e Innovación of the Spanish Government (grant PID2021-125051OB-I00) and by the Regional Government of Madrid of Spain (grant TEC 2024/COM-322).

%
%
%
%

\end{document}